\documentclass[letterpaper, 10 pt, conference]{ieeeconf}  

\IEEEoverridecommandlockouts                              

\usepackage{amsmath,amssymb,amsfonts}
\usepackage{algorithmic}
\usepackage{graphicx}
\usepackage{textcomp}
\usepackage{xcolor}
\def\BibTeX{{\rm B\kern-.05em{\sc i\kern-.025em b}\kern-.08em
    T\kern-.1667em\lower.7ex\hbox{E}\kern-.125emX}}
    
\usepackage{mathtools}
\usepackage{amsmath}
\usepackage{latexsym}
\usepackage{bm}
\usepackage{xcolor}
\usepackage{amssymb}
\usepackage{multicol}
\usepackage{pifont}
\usepackage{lipsum}

\usepackage{float}
\usepackage{booktabs}
\usepackage{multirow}
\newcommand{\myeqref}[1]{(\ref{#1})}

\title{\LARGE \bf
Highly Efficient Observation Process Based on FFT Filtering for Robot Swarm Collaborative Navigation in Unknown Environments*
}

\author{Chenxi Li$^{1}$, Weining Lu$^{2}$, Zhihao Ma$^{3}$, Litong Meng$^{1}$, Bin Liang$^{1}$
\thanks{*This work is supported by the National Natural Science Foundation of China (Grant Nos. 92248304).}
\thanks{$^{1}$Department of Automation, Tsinghua University, Beijing, 100084, China. Email:
        {\tt\small lcx22@mails.tsinghua.edu.cn}, {\tt\small menglt@ieee.org}, {\tt\small liangbin@tsinghua.edu.cn}}%
\thanks{$^{2}$Beijing National Research Center for Information Science and Technology, Tsinghua University, Beijing, 100084, China. Email:
        {\tt\small luwn@tsinghua.edu.cn}}%
\thanks{$^{3}$Department of Mechanical Engineering, Sun Yat-sen University, Shenzhen, Guangdong, 518107, China. Email:
        {\tt\small mazhh23@mail2.sysu.edu.cn}}%
}

\begin{document}
\maketitle
\thispagestyle{empty}
\pagestyle{empty}

\begin{abstract}
Collaborative path planning for robot swarms in complex, unknown environments without external positioning is a challenging problem. This requires robots to find safe directions based on real-time environmental observations, and to efficiently transfer and fuse these observations within the swarm. This study presents a filtering method based on Fast Fourier Transform (FFT) to address these two issues. We treat sensors' environmental observations as a digital sampling process. Then, we design two different types of filters for safe direction extraction, as well as for the compression and reconstruction of environmental data. The reconstructed data is mapped to probabilistic domain, achieving efficient fusion of swarm observations and planning decision. The computation time is only on the order of microseconds, and the transmission data in communication systems is in bit-level. The performance of our algorithm in sensor data processing was validated in real world experiments, and the effectiveness in swarm path optimization was demonstrated through extensive simulations.
\end{abstract}

\section{Introduction}
\label{sec:Introduction}
Robot swarms can replace humans in unknown, remote, or dangerous locations, saving manpower and costs. Autonomous navigation of robot swarms in unknown environments without external localization is challenging. Due to the randomness and complexity of the environment, robots need to process the current sensor data in real-time and update the planned path to avoid collisions with obstacles. Additionally, the target point is usually beyond the observation range of the swarm, so robots can only navigate based on local observation. For example, when encountering obstacles or walls, robots need to decide the direction of avoidance or boundary following based on current observations. However, due to limited field of view or obstruction by obstacles, it is difficult for a single robot to discern whether there are other obstacles behind the current one, which can easily lead to detours or saddle points in planning. In contrast, robot swarms can share observations of the environment to expand the field of view within the swarm, thereby achieving more efficient path planning.

Collaborative path planning for robot swarms faces significant challenges, such as real-time processing of observation data, as well as compression, transmission and reconstruction for environmental features with efficient computing. First, robots need to process sensor data in real-time. The environmental observations obtained by commonly used LiDARs or depth cameras are large amounts of high-dimensional data that cannot be directly used for path planning. A method is required to compute these sensor data to obtain a safe advancing direction for the robot. However, the computational performance of the processors equipped on commonly used small robots or UAVs is limited due to their size and battery capacity constraints. This requires the algorithm to have as low computational cost and processing latency as possible. Second, robots within the swarm need to be able to transmit and fuse observation data for path planning. Due to the large amount of raw sensor data, it requires high demands on the bandwidth and stability of the wireless communication system. For environments with limited communication capability such as in the wild, the transmission of raw sensor data directly may cost too much latency. Therefore, compression and reconstruction are required for the current observation. Considering the limitations of robot processor performance, this part of the algorithm also needs to have low computational complexity.

\begin{figure}[t]
  \centering
  \includegraphics[width=\linewidth,keepaspectratio]{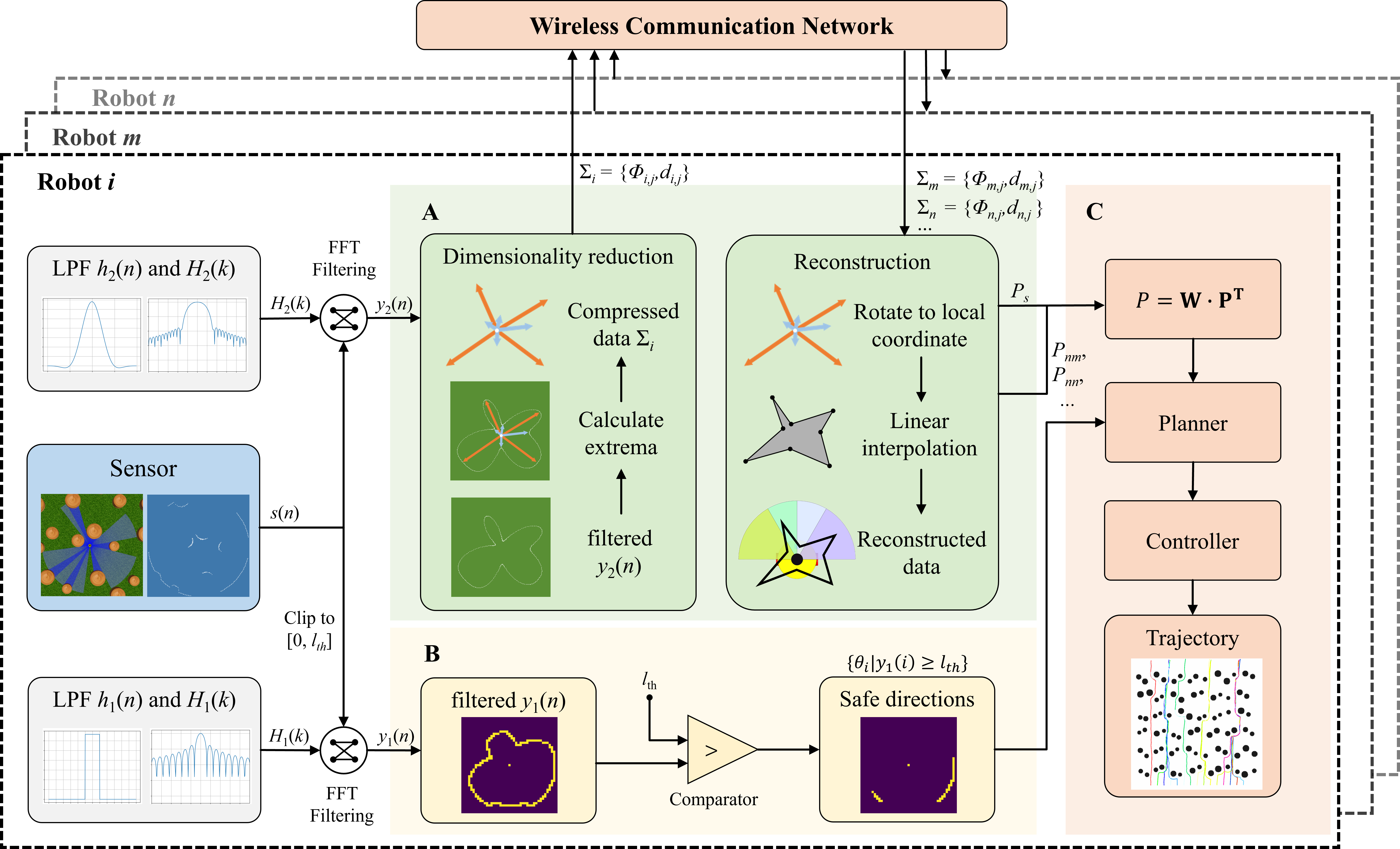}
  \caption{Architecture of our proposed method, which comprises sensors, filters, and 3 modules. (A) Dimension reduction and reconstruction. (B) Safety directions extraction from sensor data. (C) Probabilistic fusion decision and planning. The filter parameters correspond to the robot dimensions, so the filter sequences can be pre-stored in the robots' ROM. The outputs of the algorithm, consisting of safe directions and probabilistic fusion decisions, serve as inputs for the planner, where the Bug planner is used as an example.}
  \label{fig:intro}
\end{figure}

To address the two issues mentioned above, we propose to use a FFT based filtering method to process the robot's sensor data. The computational complexity of FFT is log-linear, which is much lower than the quadratic level of convolution operations. Moreover, FFT can be implemented using lightweight circuits, making it easy to deploy on various types of robots. The core of this method is the design of the filter. We designed digital low-pass filters (LPFs) to solve the two problems by combining the filter parameters with the robot's physical dimensions and path planning requirements. Then, the fast filtering is realized by FFT, where the extraction of robot's safety directions, the dimension reduction, and the reconstruction of sensor data are achieved at very low computational costs. After that, We map the robot's observations of the environment to the probability domain, thereby achieving real-time fusion of swarm observations and optimizing the robot's execution path. The architecture of our method is shown in Figure \ref{fig:intro}. In summary, our contributions are as follows:

\begin{itemize}
    \item We design a LPF according to the robot's physical dimensions. The filter can efficiently extract the safety directions from sensor data through FFT filtering. The proposed method is available for both 2D and 3D scenes, and the computation time is only on the order of microseconds. 
    \item We propose a method for fast dimension reduction and reconstruction of robot observations. It can present open space and obstacle orientation features of the robot's observation environment with bit-level data.
    \item We propose a probability-based fusion decision method. Simulation results demonstrate the effectiveness of this approach in robot swarm path optimization.
\end{itemize}

\section{Related Work}
\label{sec:related work}
For the navigation task of robot swarms in unknown environments, some existing works have solved the control problem, such as the Bug algorithms \cite{Dynamic1986,comparative2019} or Model Predictive Control (MPC) \cite{LearningB2020}. Our work is more concerned with problems at the sensing and decision levels. The aim of this paper is to address two problems mentioned above from the interdisciplinary field of digital signal filtering, which are fast extraction of safe directions from sensor data, and swarm observation fusion with low computational and communication costs.

\subsection{Processing of Robot Environmental Observations}
Robots usually cannot directly obtain safe advancing directions from the raw sensor data. Therefore, it is necessary to process the data to obtain a form that can be used to guide the path planning. Existing works can be divided into two categories based on whether an environmental map is constructed.

There are many well-established studies for creating environmental maps. For example, Simultaneous Localization and Mapping (SLAM) \cite{FastSLAM2002,VINS2018} achieves environmental reconstruction by extracting feature points in the environment and searches for paths that exist in the reconstructed map. However, this method consumes a large amount of computation and storage \cite{Research2022}. To reduce the computation cost, \cite{UFOMap2020} and \cite{UFOExplorer2022} propose to use an undirected graph and maintain a dense graph-based planning structure for each map update, thus reducing the cost of planning. \cite{EGO-Planner2021} further reduced the amount of computation by only focusing on obstacles that collide with the existing planned trajectory, so online planning is achieved and has been tested on drones. However, these methods require a large amount of time and computing power for map reconstruction before planning, and also pose significant challenges to the performance of inertial sensors and batteries. Therefore, the application scenarios are limited to small-scale environments and are not suitable for rapid search applications.

Another kind of planning methods does not require the construction of environmental maps. Instead, the planning is based on the robot's current observations. For example, some approaches establish Potential Fields (PF) \cite{Realtime1985,Mobile2022} from observed obstacle points, and guide the robot to move along the direction of the field gradient. However, this method can easily encounter saddle points in unknown environments, leading to slow progress or entering a dead end \cite{Predictive2021}. Our method is similar to frontier-based navigation \cite{frontier-based1997}. This type of approaches search for open spaces at the boundary of sensor observations and use them as optional directions for path planning, such as the Admissible Gap (AG) \cite{Admissible2018} and the Gaussian Process frontier (GP) \cite{GP-Frontier2023}. Distinct from these methods, we process sensor data through digital filtering and model the robot size constraints in filter parameters, which is faster to process with lower computation costs, and can handle data with different observation dimensions.

\subsection{Robot Swarm Collaborative Navigation}
A significant body of recent works focused on robot swarm navigation, extending single-robot methods to multi-robot systems. For example, \cite{Formation2015} extended the Bug algorithm to robot formations using the leader-follower control strategy. \cite{Minimal2019} proposed the Swarm Gradient Bug Algorithm (SGBA) based on the Received Signal Strength Intensity (RSSI) of external beacons, improving the exploration efficiency of swarms in unknown environments. \cite{Swarm2022} extended \cite{EGO-Planner2021} to micro flying robot swarms, achieving traversal and tracking in outdoor environments by joint optimization of spatial-temporal trajectories with wireless broadcasting trajectories. \cite{Predictive2021} improved the saddle point problem of PF methods and proposed the Nonlinear Model Predictive Control (NMPC), using a central computing node to optimize swarm trajectories, achieving navigation in cluttered environments without collisions.

Although previous works provide new ideas for collaborative navigation of robot swarms, currently no work can effectively utilize the collective observations of the swarm to the environment. Due to the vast amount of raw sensor data and the limitations of communication system bandwidth, achieving real-time transmission and fusion of swarm observations remains a challenge.

\section{Methodology}
\label{sec:Methods}
\subsection{Robot Protective Distance Model}
\label{subsec:protective distance model}
\begin{figure}[ht]
  \centering
  \includegraphics[width=\linewidth,keepaspectratio]{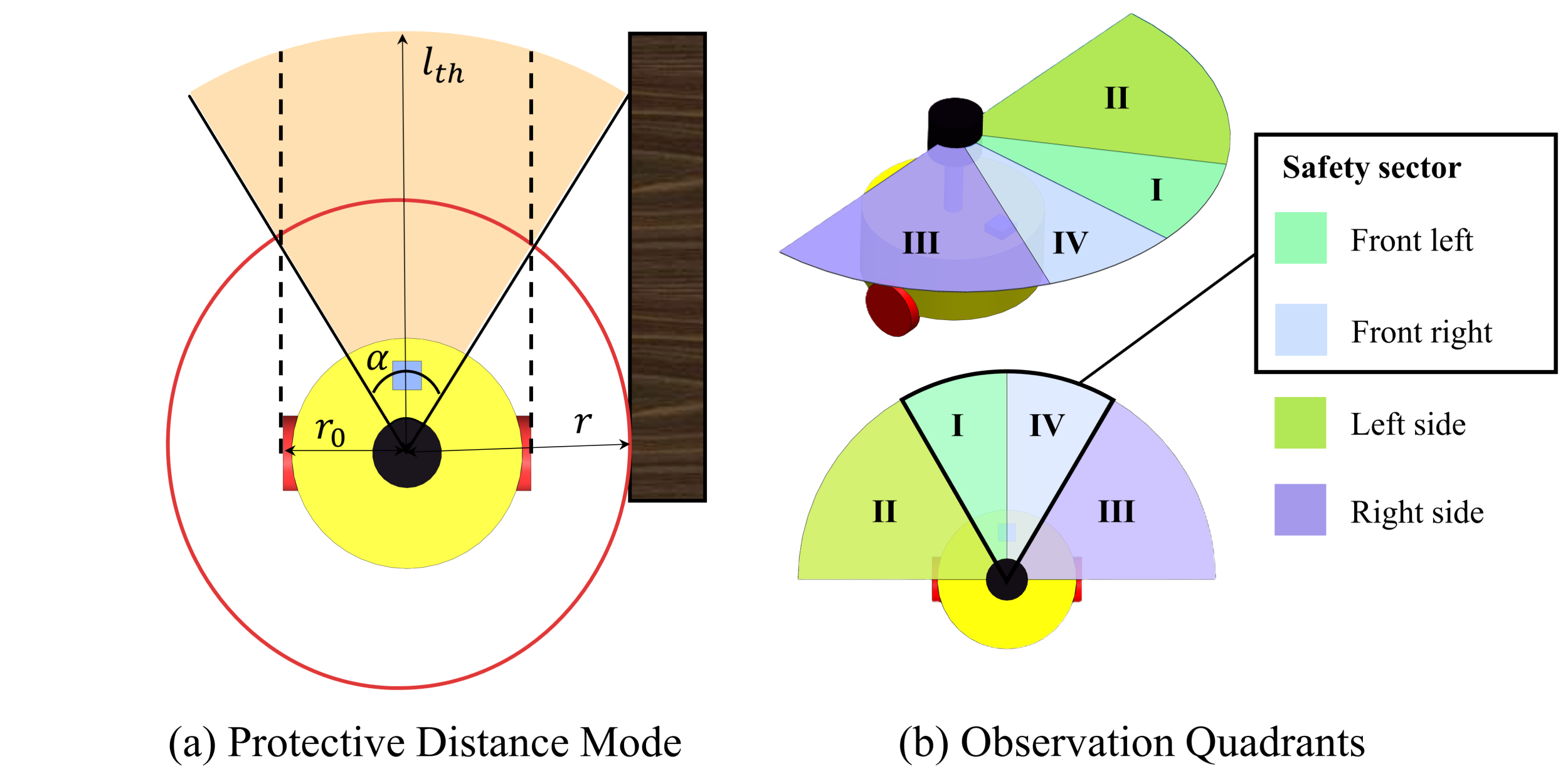}
  \caption{Protective Distance Model. (a) Geometric relationship between the robot collision radius $r_0$, protection radius $r$ and planning distance $l_{th}$. (b) The observation of the robot is divided into 4 quadrants based on left and right, as well as whether it is within the safety sector.}
  \label{fig:protect}
\end{figure}

A robot operating in the real world is not a point mass but has a volume that need to be accounted for. Let its lateral collision radius be $r_0$. In order to be compatible with robots of different shapes, we design a protective distance model as shown in Fig. \ref{fig:protect}a. To prevent collisions, we preserve a safety margin between robots and other objects in path planning, which is called the protective radius $r$. The choice of the parameter should consider the robot’s outer circle radius and the flexibility of the lateral motion. A larger $r$ allows the robot to plan safely for a greater distance each time. Let the planned distance be $l_{th}$. When the robot's safety field of view is greater than $\alpha$, it ensures the robot will not collide within the distance of $(l_{th}-r_0)$. This area is called safty sector. $l_{th}$ and $\alpha$ can be calculated through geometric relations:
\begin{equation}
l_{th} = \frac{r}{\cos{\left( \frac{\pi - \alpha}{2}\right)}} = \frac{r^2}{r_0}
\label{eq:lth}
\end{equation}
\begin{equation}
\alpha = \pi - 2\arccos{\left( \frac{r_0}{r}\right)}
\label{eq:alpha}
\end{equation}

Generally, the robot only needs to decide its direction of advancement based on its forward observations. We divide the robot's observation area into four quadrants: front-left (I), left-side (II), right-side (III), and front-right (IV), as shown in Fig. \ref{fig:protect}b. The front-left and front-right quadrants are divided by the robot's orientation, splitting the safety sector into two halves. The left-side and right-side quadrants focus on the environment in the forward side areas outside the safety sector.

\subsection{Filter Parameters and Filtering Process}
\label{subsec:filter parameters and filtering process}
The observation by robots can be considered as a form of digital sampling. Let the robot's field of view be $\Theta$, with a fixed resolution, and the number of sampling points be $M$. So the digital sampling frequency is $f_s = M$. Then, the 2D circular observation can be represented as a distance sequence $s(n)$. We use the maximum observation distance $R$ for normalization:
\begin{equation}
s(n) = s[\bm{\mathbf{\theta}}] = \frac{\mathbf{d}}{R}, n=0,1,\dots,M-1
\label{eq:sensor data}
\end{equation}
where \( \bm{\theta} = [\theta_1, \theta_2, \ldots, \theta_M] \) are the values of angles uniformly sampled with respect to the current orientation of the robot, and \( \mathbf{d} = [d_1, d_2, \ldots, d_M] \) are the corresponding observed distances. According to the model in Section \ref{subsec:protective distance model}, the width of a space is sufficient for the robot to pass through only if its safety sector angle is larger than $\alpha$ in the direction. From the perspective of digital signals, the wider environmental space, the lower the digital frequency of $s(n)$. Conversely, if the space is narrow, the digital frequency of $s(n)$ will be high. Therefore, digital low-pass filtering can be used to select directions that meet the width requirements of the space for robot path planning. The cutoff frequency in analog domain \(f_0\) is the robot's field of view \(\Theta\) normalized by the safety sector angle \(\alpha\), given by:
\begin{equation}
f_0 = \frac{\Theta}{\alpha}
\label{eq:f_0}
\end{equation}
so the cutoff frequency of the LPF in digital domain should be:
\begin{equation}
f_c = \frac{f_0}{f_s}=\frac{\Theta}{M\alpha}
\label{eq:f_c}
\end{equation}

According to Fourier transform theory, the circular convolution in the signal's time domain is equivalent to the product of signals in the discrete Fourier domain. Therefore, in Sections \ref{subsec:extracting safety Directions} and \ref{subsec:dimension reduction and reconstruction}, we designed the time-domain of the filters \(h_1(n)\) and \(h_2(n)\), transformed them to frequency domain and get \(H_1(k)\) and \(H_2(k)\). They are to be stored in the robot's ROM. When processing the sensor data \(s(n)\), simply transform it to the frequency domain \(S(k)\) using FFT, and then complete the multiplication operation in frequency domain to achieve filtering process. The filtered sensory data can be obtained by taking the principal value sequence of the output signal. Let the number of points in the FFT be $N_0$. Compared to traditional convolution methods, FFT requires only $(\frac{N_0}{2} \log_2 N_0)$ multiplications and $(N_0 \log_2 N_0)$ additions in the field of complex numbers \cite{Algorithm1965}. Furthermore, FFT does not require pre-training, and can be implemented with digital circuits. This significantly reduces the hardware requirements of the algorithm on the processor and greatly reduced the processing delay.

To avoid signal aliasing, the number of points \(N_0\) for the FFT needs to meet certain requirements. Let the filter points be $N_1$ and $N_2$ respectively. First, to satisfy the constraints of the Nyquist theorem, it is required that \(N_0 > 2\max(M, N_1, N_2)\). Second, to avoid signal front-to-end aliasing, it is necessary to satisfy \(N_0 > \max(M + N_1 - 1, M + N_2 - 1)\). Finally, to balance computational efficiency, \(N_0\) should be chosen as the smallest power of 2 that meets the above conditions.

The FFT process can be extended to 2D signals by first processing rows and then columns. Therefore, we can represent the 3D spherical distance observation in the form of a binary function of spherical coordinates:
\begin{equation}
s(u,v) = s[\bm{\mathbf{\phi,\theta}}] = \frac{\mathbf{D}}{R}
\label{eq:3D sensor data}
\end{equation}
where \( \bm{\mathbf{\phi,\theta}} \) are the values of angles uniformly sampled in spherical coordinate system, and $\mathbf{D}$ is the corresponding distance matrix. This allows our approach to handle 3D environmental sensory data such as those observed by UAVs, and thus to adapt to navigation scenarios in different dimensions.

\begin{figure}[ht]
  \centering
  \includegraphics[width=\linewidth,keepaspectratio]{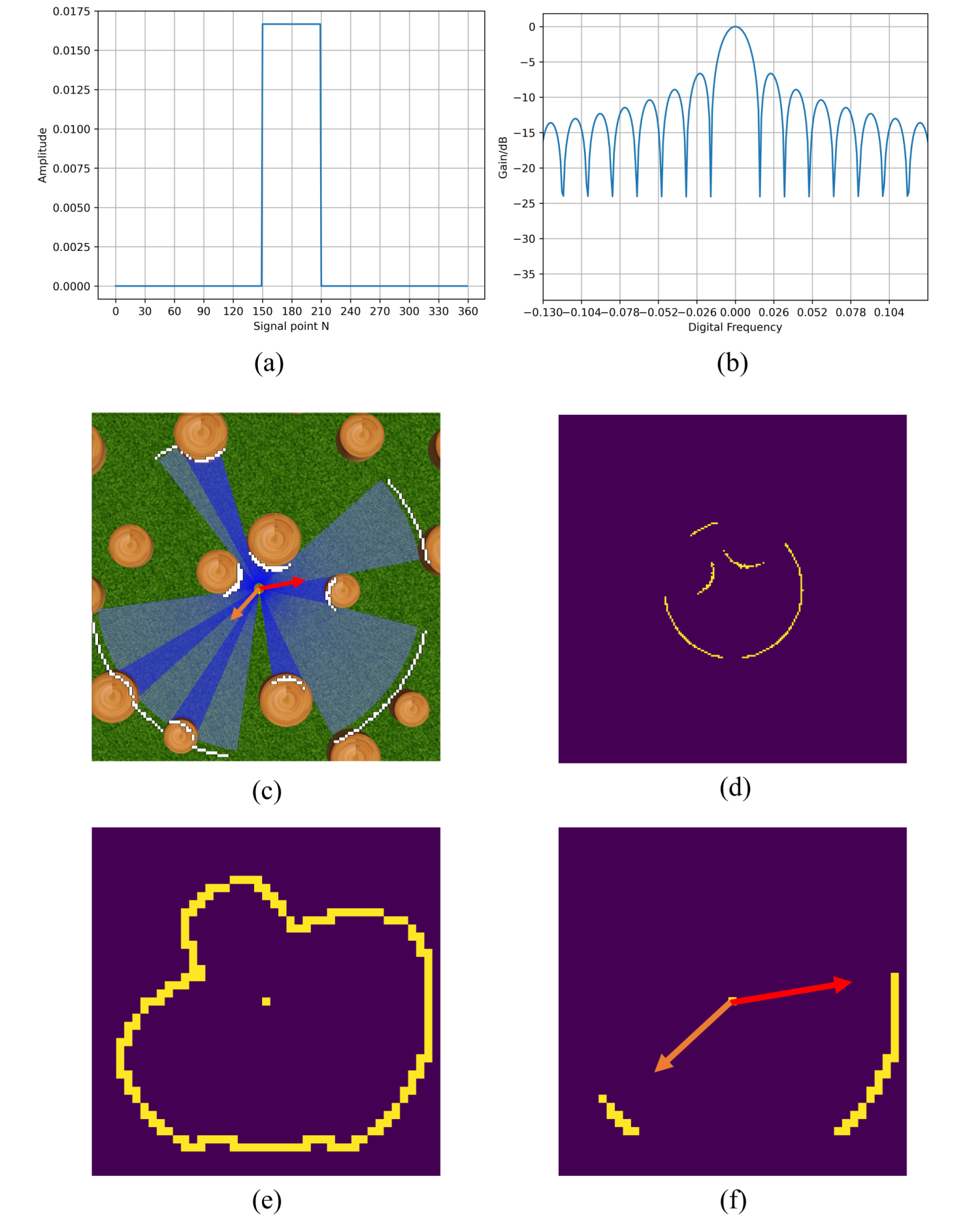}
  \caption{Illustrative example of extracting safety directions from sensor data. Simulation parameters are presented in Section \ref{subsec:resultes-planning}. (a) Filter $h_1$ in the time domain. The width of the rectangular window $T_c$ corresponds to the robot's safe sector angle $\alpha$. (b) Filter $H_1$ in the frequency domain. The FFT points $N_0=1024$. The digital cutoff frequency is $f_c=0.013$, at which the filter gain is -3dB. (c) The robot's sensor data points overlaid on the picture of the working environment. The two arrows in the figure represent the optional forward directions towards the target direction. (d) Truncated observation data using the planning distance of $l_{th}=0.6m$. (e) The data after filtering in normalized scale. (f) The safe direction determined using condition \myeqref{eq:y1 for safe}. The two arrows in the figure correspond to those in (c).}
  \label{fig:h1}
\end{figure}

\subsection{Extracting Safety Directions from Sensor Data}
\label{subsec:extracting safety Directions}
In this section, we design a filter and an algorithm to rapidly extract safe forward directions from the observed data through FFT filtering. For \(\forall{\theta_i}\) in \(\bm{\theta}\), the necessary and sufficient condition for this direction to be safe is: all corresponding observed distances $s(n)$ are greater than the normalized planning distance \((l_{th} / R)\) within the safety sector width \(\alpha\) centered at \(\theta_i\). This width in digital domain is $T_c = \lceil \frac{1}{f_c} \rceil$. So the condition can be expressed as:
\begin{multline}
    \theta_i \text{ is safe} \iff \text{for } n = (i-T_c) \text{ to } (i+T_c),\\
    s(n) \geq \frac{l_{th}}{R}, T_c < i < M - T_c
\label{eq:necessary and sufficient condition for safe}
\end{multline}

Hence, we truncate \(s(n)\) at a maximum value of \((l_{th} / R)\), and then use a rectangular window with a width of \(T_c\) as the time domain of the filter:
\begin{equation}
h_1(n) = \frac{1}{T_c} R_{T_c}(n - \tau_1), n=0,1,\dots,N_1-1
\label{eq:h1}
\end{equation}
where \(N_1\) is the length of the filter, and \(\tau_1 = \frac{N_1 - 1}{2}\) is the group delay. To meet the conditions for linear phase filtering, \(N_1\) should be an odd number. \(R_{T_c}\) is the rectangular window sequence with a width of \(T_c\). To save computational cost while ensuring phase and accuracy match, we set \(N_1 = M\). The frequency domain \(H_1(k)\) is obtained by applying FFT to \(h_1(n)\). The time and frequency domain of the filter and filtering process are shown in Figure \ref{fig:h1}. Although the filter is not a typical LPF, it can be considered as a sliding window used to determine whether the distance observations within the window meet the requirements. Let the filtered signal sequence be \(y_1(n)\). Therefore, the condition of \myeqref{eq:necessary and sufficient condition for safe} can be transformed for filtered sensor data:
\begin{equation}
    \theta_i \text{ is safe} \iff y_1(i) \geq \frac{l_{th}}{R}
\label{eq:y1 for safe}
\end{equation}
In robot path planning process, applying the filter \myeqref{eq:h1} can quickly extract the safe direction at the scale of planning distance \(l_{th}\), and serve as the input for the planner.

\begin{figure*}[ht]
  \centering
  \includegraphics[width=0.9\textwidth,keepaspectratio]{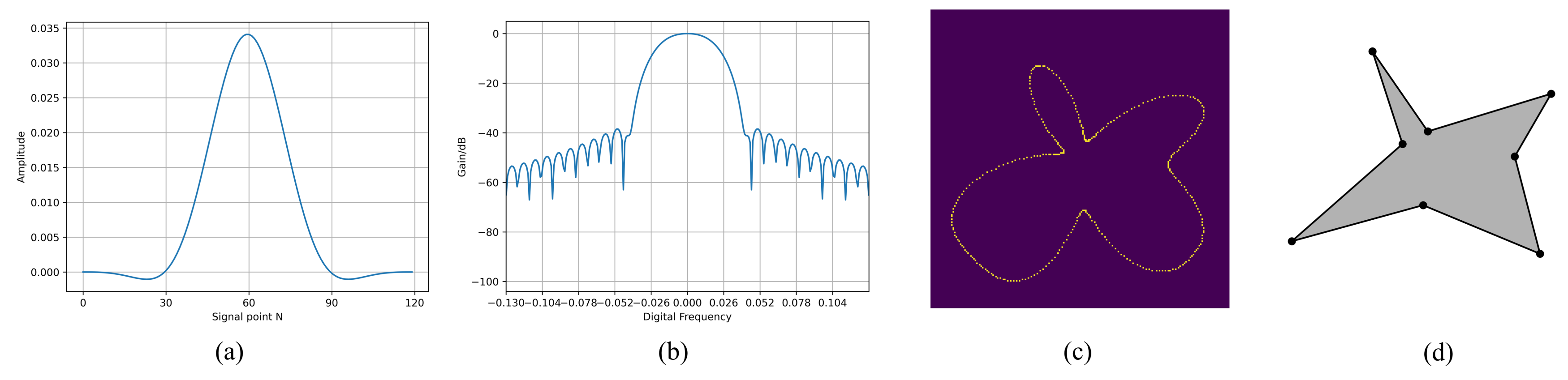}
  \caption{Filter design for dimension reduction and reconstruction of sensor data. The environment and the robot observations are the same as those in Figure \ref{fig:h1}c. (a) Filter $h_2$ in the time domain. The main lobe width of the filter corresponds to the robot's safe sector angle $\alpha$, and the sidelobe decays rapidly. (b) Filter $H_2$ in the frequency domain. The filter gain is -3dB at digital cutoff frequency $f_c=0.013$. (c) Filtered sensor data. Dimension reduction can be achieved by finding the extrema, as described in \myeqref{eq:extrema}. The compressed data for transmission in normalized scale is $\Sigma_0=$ \{(26, 0.42), (67, 0.074), (125, 0.598), (183, 0.208), (229, 0.606), (267, 0.358), (294, 0.542), (342, 0.08)\}. (d) The reconstructed observation based on $\Sigma_0$.
  }
  \label{fig:h2}
\end{figure*}

\subsection{Dimension Reduction and Reconstruction of Sensor Data}
\label{subsec:dimension reduction and reconstruction}
In this section, we design a LPF based on the robot's physical dimensions described in Section \ref{subsec:protective distance model}. Key features of the robot's surrounding environment, including open space and obstacle orientation, are quickly extracted from sensor data by filtering. Then, the features are utilized to achieve observation compression and dimension reduction, thereby facilitating low-cost data transmission and fusion within the swarm.

\paragraph{Filter design} We use window function method \cite{On1978} to design the digital filter. Let the number of filter points be \( N_2 \). Firstly, we design an ideal linear-phase LPF with a cutoff frequency of \( f_c \). Then, truncate it using a window function $w_{N_2}(n)$, the time domain of the filter can be obtained:
\begin{equation}
h_2(n) = 2f_c \text{sinc} \left(2f_c(n-\tau_2) \right)w_{N_2}(n)
\label{eq:h2}
\end{equation}
where \( \tau_2 = \frac{N_2 - 1}{2} \), and $\text{sinc}(x)=\frac{sin(\pi x)}{\pi x}$. The choice of window function affects the time-frequency resolution and the transition bandwidth of the filter. In fact, we aim to extract the orientations of open spaces and obstacles from the time domain of the filtered signal sequence \(y_2(n)\). Therefore, the resolution of the window function is more important compared to the transition bandwidth. In addition, to smooth the boundaries of the filtered signal and reduce clutter interference, the window function needs to have a higher stopband attenuation and fast decay sidelobe. Hence, we choose the Blackman window function \cite{On1978}. The time-domain expression is $w_{N_2}(n) = 0.42 - 0.5 \cos(\frac{2\pi n}{N_2-1}) + 0.08 \cos(\frac{4\pi n}{N_2-1})$. The time and frequency domain of the filter are shown in Figure \ref{fig:h2}. Since the sum of the values in the time domain sequence \(h_2(n)\) is 1, \(y_2(n)\) can be considered as the weighted average of the environmental depth within the field of view.

\paragraph{Dimension reduction} The main lobe width of the filter \(h_2(n)\) corresponds to the robot's the safety sector angle \(\alpha\) in physical space. Thus, the numerical result of \(y_2(n)\) represents the passability of each direction \(\bm{\theta}\). The larger the value of \(y_2(n)\), the larger the open space in the corresponding direction. By finding the extrema of \(y_2(n)\), we can locate the positions of open spaces and obstacles at the scale of \(\alpha\). These are the key features of the robot's surroundings observation, and we use them to achieve dimension reduction of the observation data. The number of maxima or minima is around the cutoff frequency $f_0$, so the amount of data after dimension reduction is very small. For robot $i$, let the number of extrema be $N_i$, then the data after reduction can be represented as:
\begin{equation}
\left \{\Sigma_i = (\phi_{i,j}, d_{i,j}) | j = 1, 2, \ldots, N_i \right \}
\label{eq:extrema}
\end{equation}
where \(\phi_{i,j}\) and \(d_{i,j}\) respectively represent the direction and distance of the extrema. If the data is in float format, then the volume is around \((16 * f_0 \)) bytes, which greatly reduces the demand on wireless communication systems and ensures real-time data transmission within swarms.

\paragraph{Reconstruction from compressed data} Robots transmit the compressed observations within the swarm over a wireless network. When a robot receives $\Sigma_i$ from neighbor robot $i$, it needs to rotate $\phi_{i,j}$ to local coordinate system to achieve observation fusion. The original data can be approximately reconstructed by circular or linear interpolation between extreme points. For simplicity, we uses linear interpolation, obtaining approximate environmental representation in polygonal forms. The details of the whole process are shown in Figure \ref{fig:intro}A and Figure \ref{fig:h2}.

\subsection{Probabilistic fusion and decision}
\label{subsec:probabilistic description}
When a robot encounters an obstacle while moving forward, it needs to choose a direction to follow the boundary of the obstacle, either left or right. We map the data reconstructed in Section \ref{subsec:dimension reduction and reconstruction} to the probability domain, so as to achieve the fusion of the robot's own observations with those received from neighbors. The direction of advance is determined based on the fused probability values.

\paragraph{Robot's own probability \( P_s \)} When an obstacle is encountered in the forward safety sector, the robot chooses a direction to bypass the obstacle based on its observation. We use a sigmoid function $\bm{\sigma}(\cdot)$ to map the robot's sensor data to the probability of turning left, given by:
\begin{equation}
P_s = \bm{\sigma}(\mathbf{W_0^{\top}} \mathbf{x_0}+b_0)
\label{eq:P_s}
\end{equation}
where $\mathbf{x_0}$ is a 4-dimensional vector, with each component representing the estimated open space area observed in the four quadrants shown in Fig \ref{fig:protect}. $b_0$ is the bias entry usually set to zero in environments without a priori. $\mathbf{W_0}$ is the parameter vector, representing the weights in different quadrants. We use the area of the linear interpolation polygon that falls within four quadrants as an approximate characterization of the robot's surrounding environment. This simplifies the model input to a four-dimensional vector \( \mathbf{x_0} \), where the value of each dimension is:
\begin{equation}
x_{0,k} = \sum_{j=0}^{n_{0,k}} \frac{1}{2} d_{0,j} d_{0,j+1} \sin{\left(\phi_{0,j+1} - \phi_{0,j}  \right)}, k=1,2,3,4
\label{eq:x_i}
\end{equation}
where \( n_{0,k} \) represents the number of extrema within the robot's \( k \)-th observation quadrant. The values of \( \phi_{0,j}, d_{0,j} \) for \( j=1 \) to \( n_{0,k} \) indicate the direction angles and distances of the extrema within the $k$-th quadrant, while for \( j=0 \) and \( j=n_{0,k}+1 \) represent the value of two quadrant boundaries. In most cases, the importance of left and right can be considered equal, so the absolute values of $\mathbf{W_0}$ are symmetric but opposite in sign:
\begin{equation}
\mathbf{W_0}=\frac{[w_s,1,-1,-w_s]^{\top}}{(1 + w_s)}
\label{eq:W_0}
\end{equation}
where \( w_s \) is the weight ratio of observations inside and outside the safety sector. We set \( w_s=1 \) in our experiments.

\paragraph{Neighboring robot's probability \( P_{i} \)} The physical connotation of \(P_{i}\) is the passable probability towards the target direction at the location of neighbor \(i\). As described in Section \ref{subsec:dimension reduction and reconstruction}, the filtered data represents the weighted average environmental depth, or the passability at the scale of \(\alpha\) within the field of view. Therefore, after receiving the observation \(\Sigma_i\) from neighbor $i$, we use the reconstructed data as input, and map it to the probability domain using the \(\bm{\sigma}(\cdot)\) function:
\begin{equation}
P_{i} = \bm{\sigma}(x_{ni}+w_n)
\label{eq:P_ni}
\end{equation}
where \(x_{ni}\) denotes the observation in safety sector towards the target direction, and \(w_n\) is the threshold. If no local minimum exists within the sector, it indicates that there are possibly no direct obstacles in the direction. Therefore, $x_{ni}$ is assigned with the local maximum value or the maximum boundary value in the sector. If a local minimum exists within the sector, it indicates the potential presence of obstacles. In such cases, we adopts a fuzzy evaluation, using the average of the minimum and maximum values for assignment:
\begin{equation}
x_{ni} = 
\begin{cases} 
\max_{k}{\left\{d_{i,k}    \right\}  } & \text{no minima}\\
\text{avg} \left(\max_{k}{\left\{d_{i,k}    \right\}  }, \min_{k}{\left\{d_{i,k}    \right\}  }  \right) & \text{minima exist}
\end{cases}
\label{eq:x_1}
\end{equation}
\(k\) is the index of the extreme values falling within the observation quadrants I and IV. As for the threshold \( w_n \), it is set to twice the normalized longitudinal scale of the robot's safety domain $(l_{th}+r)$, corrected with the robot radius $r_0$. We have \( w_n = \left(-\frac{2(l_{th}+r)+r_0}{R} \right) \). Hence, a higher value of \( P_{i} \) is output only when there is a locally passable way in the neighbor's position.

\paragraph{Fusion and decision process} After the calculation of probabilities for itself and its neighbors, the robot needs to fuse in the probability domain to choose a direction with fewer obstacles or more open space, thus optimizing its own execution path. Due to the limitation of the field of view, robots only fuses the neighbors within its observation quadrants. Let the number of neighbors located in the left quadrants I and II be \(n_l\), and the number of neighbors in the right quadrants III and IV be \(n_r\), then the fusion process can be represented as:
\begin{equation}
P = w_0 P_s + \sum_{i=1}^{n_l} w_i P_{i} + \sum_{i=n_l+1}^{n_l+n_r} w_i (1-P_{i}) = \mathbf{W} \cdot \mathbf{P} ^ {\top}
\label{eq:P}
\end{equation}
where \(P\) is the probability of turning left after fusion. \(w_0\) is the weight of the robot's own observation, and \(w_i\) is the weight of the neighbors' observations. The fusion process can be written in matrix form. \(\mathbf{W}=[w_0, w_1, \dots, w_{n_l+n_r}]\) is the weight matrix, where the sum of the matrix elements is 1. \(\mathbf{P}=[P_s, P_{1}, \dots, P_{n_l}, (1-P_{n_l+1}), \dots, (1-P_{n_l+n_r})]\) is the probability matrix. To ensure the effectiveness of the fusion, neighbors with overlapping protective radius and those too far away were excluded, and we assign their weights to 0. The remaining neighbors were assigned the same weight. Since the neighbor observations received are biased relative to the robot's own position, the neighbors' weights \(w_i\) should be less \(w_0\). In simulation tests, we set \(w_i = \frac{1}{3}w_0\).

\section{Performance Evaluation}
\label{sec:Results}
In this section, we first tested the algorithm's performance in processing sensor data in the real world, as well as applicability in 3D observation in simulation. Then, we tested the planning performance in complex unknown environments using a swarm of 15 robots on Gazebo platform, demonstrating the effectiveness of our proposed probabilistic fusion decision method.

\subsection{Sensor Data Processing}
\label{subsec:resultes-sensor}
\begin{figure*}[t]
  \centering
  \includegraphics[width=0.9\textwidth,keepaspectratio]{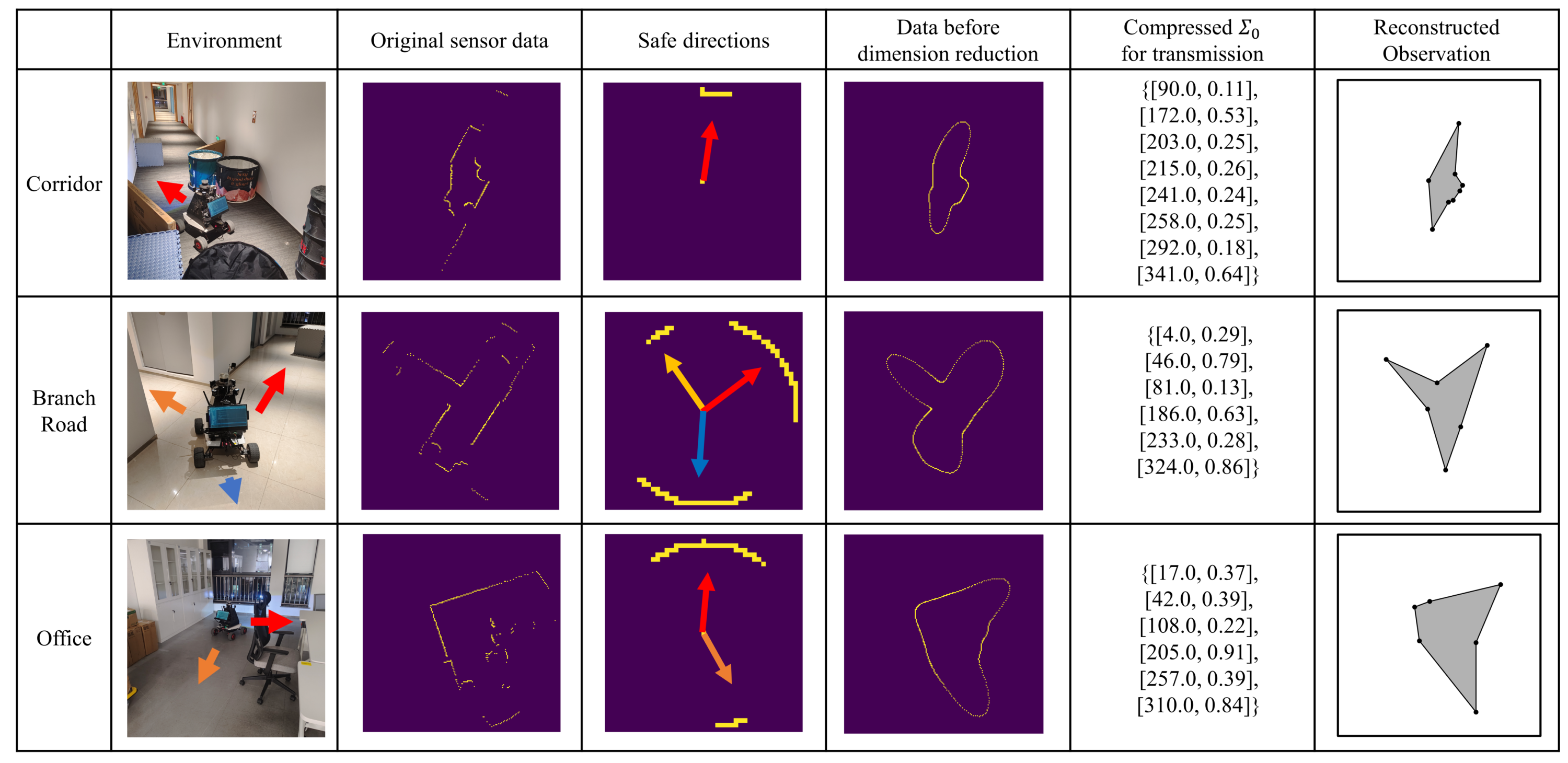}
  \caption{Sensor data processing in real world experiments. The arrows in the figure indicate open spaces in the environment that are able to pass through.}
  \label{fig:sensor-data}
\end{figure*}

The simulation tests of our proposed algorithm have been used as examples of filtering process in Section \ref{sec:Methods}, as shown in Figure \ref{fig:h1} and Figure \ref{fig:h2}. Here, we conduct tests in the real world. A wheeled robot equipped with an Ouster-OS0 LiDAR and an 8-core ARMv8 CPU was used to collect and process real world sensor data. The robot's size is $625\rm{mm} \times 585\rm{mm}$, with a lateral collision distance of approximately $r_0=0.29\rm{m}$ and a circumscribed circle radius of about $0.43\rm{m}$. The robot in real-world experiments is larger than the simulation one (detailed in Section \ref{subsec:resultes-planning}), so we set the protection radius to $r=0.7\rm{m}$. According to \eqref{eq:lth} and \eqref{eq:alpha}, we can obtain $l_{th}=1.69\rm{m}$ and $\alpha=49^\circ$. We tested the sensor data processing results in cluttered corridor, branch road and office environments, as shown in Figure \ref{fig:sensor-data}.

\begin{figure}[ht]
  \centering
  \includegraphics[width=0.9\linewidth,keepaspectratio]{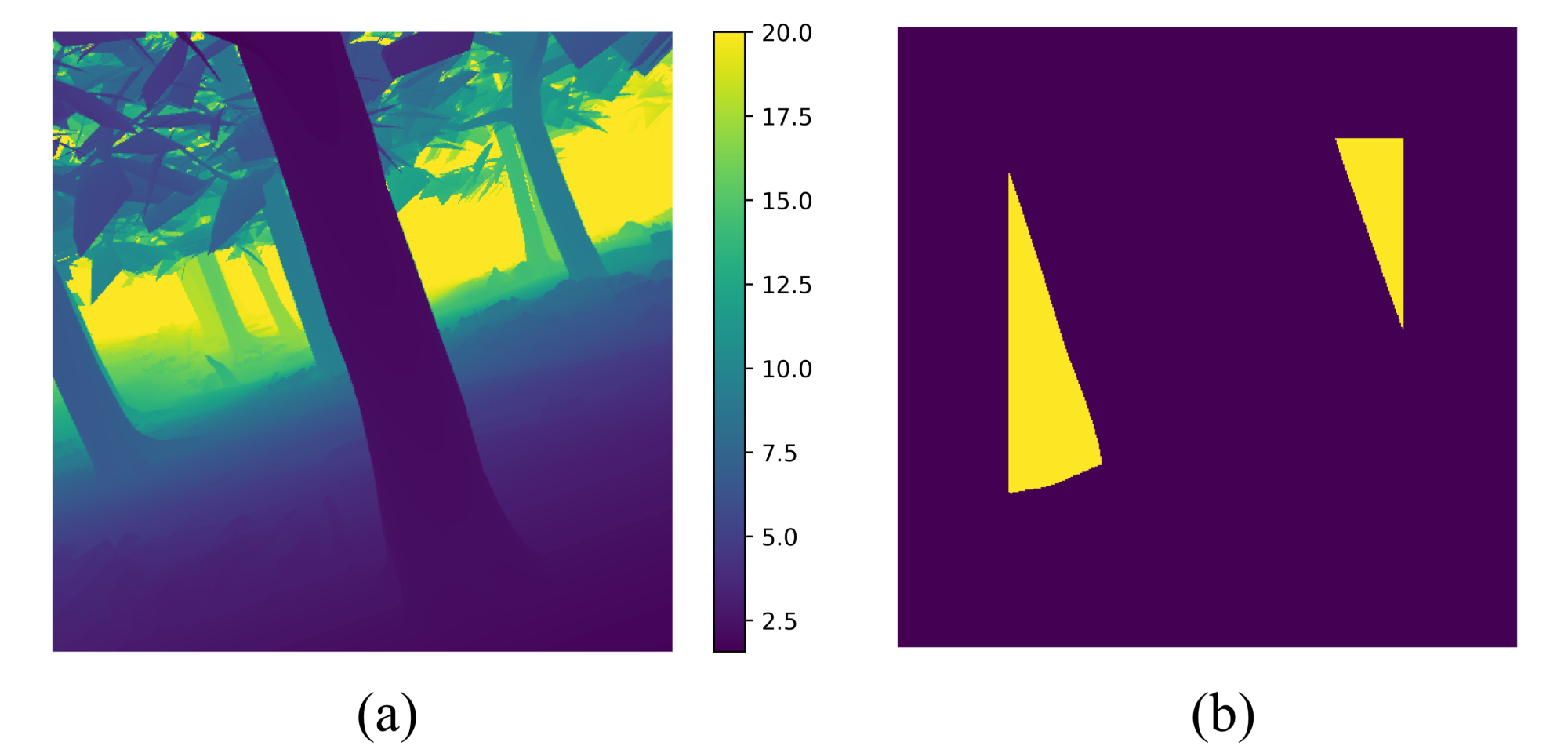}
  \caption{Our algorithm processes 3D observations. (a) Depth observation of the forest. The color bar represents the depth value in meters. (b) Safe direction domain extracted by filtering.}
  \label{fig:3D cases}
\end{figure}

In terms of safe direction extraction, our algorithm can find the range of directions that satisfy robot size and planning distance requirements, and represent them with angles relative to the robot's orientation. This can be directly used as input for planning and control algorithms. In terms of sensor data dimension reduction and reconstruction, our algorithm can use only a few data points to approximately represent the robot's observation. The amount of data after dimension reduction is only a few dozen bytes in float format. This ensures the swarm observations transmitted in real time, and allows our algorithm to be applied in environments with limited communications. We also tested the processing latency of the algorithm. Under tested hardware conditions, the average processing latency is only 13$\rm{\mu s}$, which is $10^3$ orders of magnitude lower than the perception latency of LIDAR. If the FFT computation is implemented with hardware circuits, the processing latency will be even lower. Therefore, our algorithm can be applied to lightweight and low power robot platforms.

We also try to extend the algorithm in the field of 3D observation. We tested in wooded environments using a quadrotor equipped with a depth camera on Unity platform. The drone width is $r=0.15\rm{m}$. We set the protection radius $r=0.65\rm{m}$, so $l_{th}=2.82m$ and $\alpha=27^{\circ}$. We use circular rotation to expand \myeqref{eq:h1} to 2D, and then use 2D FFT to perform the computation. The computation results for the safe passage direction are shown in Fig. \ref{fig:3D cases}. It can be seen that our method can be extended to 3D application scenarios and deployed on flying robots. This area has the prospect of deeper research.

\subsection{Swarm Collaborative Path Planning Performance}
\label{subsec:resultes-planning}
\begin{figure}[ht]
  \centering
  \includegraphics[width=0.9\linewidth,keepaspectratio]{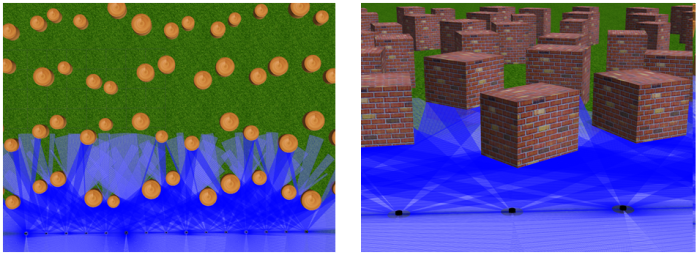}
  \caption{Simulation platform for swarm navigation tests, including dense forest and rocky ruin environments respectively.}
  \label{fig:sim-platform}
\end{figure}

In order to demonstrate the effectiveness of our proposed probabilistic fusion decision method, we conducted swarm traversal tests in complex environments without external positioning or prior map information. We tested with swarms of 15 robots in 2 kinds of environments, dense forest and rocky ruins, using Gazebo platform in Figure \ref{fig:sim-platform}. We randomly generate 20 maps for each environment. The size of the environment were $20\rm{m} \times 20\rm{m}$, and the density of trees or rocks is about 1 per 5 square meters. Robots radius $r_0=0.15\rm{m}$, protection radius $r=0.3\rm{m}$, and thus $l_{th}=0.6\rm{m}, \alpha=60^{\circ}$. Robots are capable of moving straight or rotating in place. The LiDAR used in simulation has an observation range of $R=3m$, the field of view is $\Theta=360^{\circ}$, and the number of sampling points is $M=360$. But there is a $15^{\circ}$ blind area in the back with a constant observation distance of 0. We use the Bug algorithm as an example to illustrate the effectiveness of our proposed fusion and decision method in planning. Our method can be integrated into other basic planning approaches, not limited to Bug planners. The safe directions are extracted by the filtering method in Section \ref{subsec:extracting safety Directions} as inputs to the planner. When the robot encounters an obstacle, we use the method in Section \ref{subsec:dimension reduction and reconstruction} to transmit the observation data within the swarm, and achieve probabilistic fusion and decision described in Section \ref{subsec:probabilistic description} to select the direction in which the robot will follow around the obstacle boundary.

\begin{figure}[ht]
  \centering
  \includegraphics[width=\linewidth,keepaspectratio]{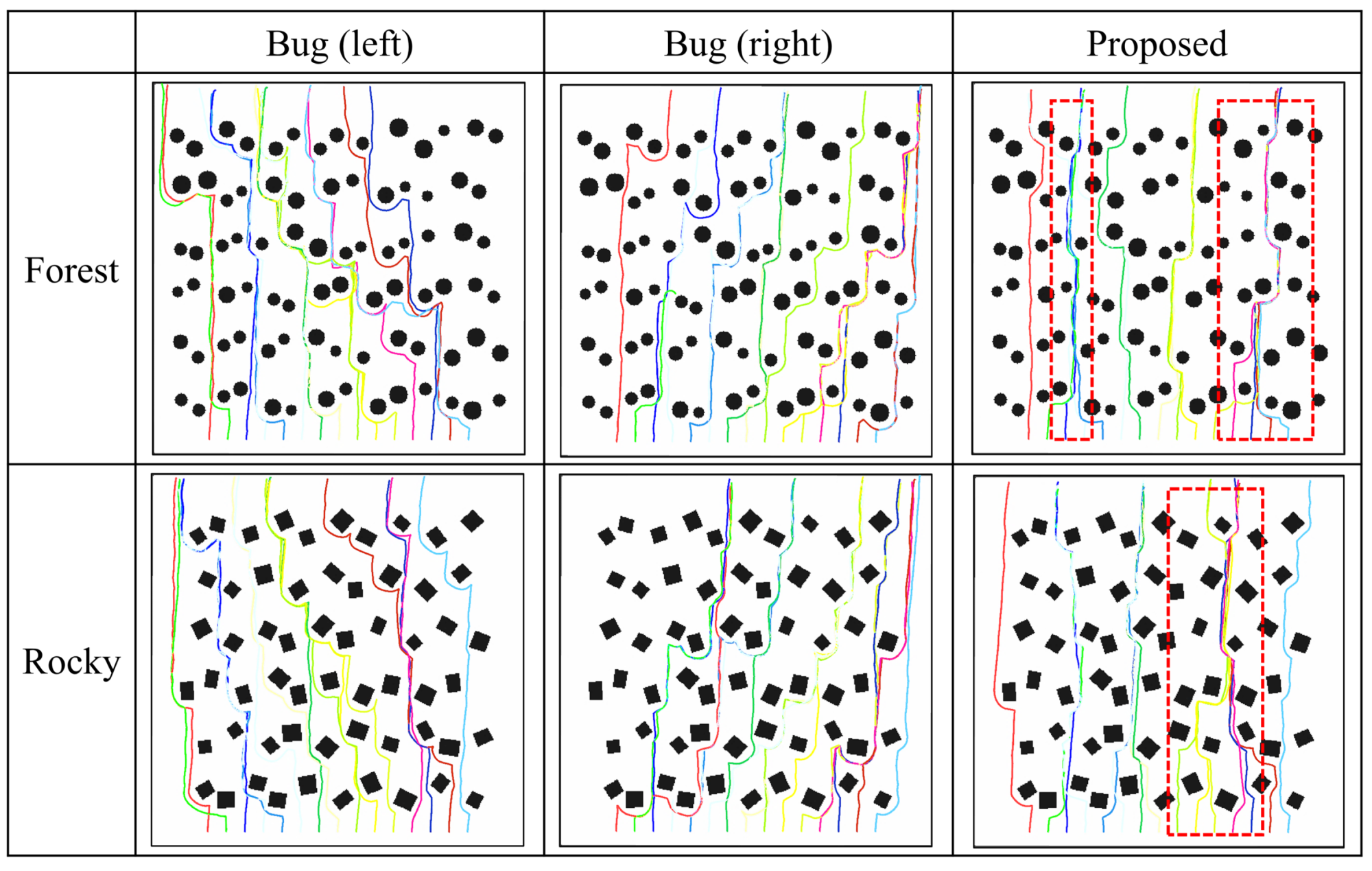}
  \caption{Performance comparison between our proposed probabilistic fusion decision method and the baseline for robot swarm planning. Here presents an example for one of the randomized maps. A fusion effect can be observed from the trajectories in the red rectangular box. It shows an effect of group aggregation and optimizes execution paths.}
  \label{fig:sim-fig}
\end{figure}

To evaluate the performance, we used the following statistical metrics: (a) Robot Arrival Rate (AR). (b) Average Path Length (APL). (c) Success weighted by inverse Path Length (SPL)\cite{Evaluation2018}, defined as the ratio of the the optimal path to actual path length weighted by the success rate. This metric comprehensively reflects the success rate and path efficiency of the robots: the closer this value is to 1, the closer the robot's execution path is to the optimal path. Here, we use the results of the A* algorithm \cite{Formal1968} as the optimal path, with a resolution of 0.05m. We compared the performance with Bug baselines as shown in Figure \ref{fig:sim-fig} and Table \ref{sim-forest}.

Experimental results demonstrate the ability of our proposed method for safe traversal of robot swarms through complex unknown environments, achieving significant improvements in terms of APL and SPL compared to the baseline. Although baseline planners manage to complete the traversal in the majority of cases, a small number of robots fail to accomplish, which mainly due to collisions occurred in dense scenarios. Furthermore, its constant decision pattern in path planning leads to a significant amount of redundant paths, as well as a noticeable deviation of the swarm's lateral position from the initial position after traversal. Different from baselines, our proposed method chooses the direction of forward and obstacle following by integrating observations from both the robot itself and its neighbors. When a robot within the swarm detects a significant open space in the target direction, it processes and transmits the observation to its surrounding neighbors using the method in Section \ref{subsec:dimension reduction and reconstruction}. Upon receiving this observation, the surrounding robots form a high fusion probability towards the sender's direction, thereby exhibiting a preference to move in that direction. Through the real-time sharing and probabilistic fusion of observations within the swarm, a mechanism is formed where robots at the front scout and those following adjust their paths based on the front robots' environmental observations.

\begin{table}[h]
\caption{Performance statistics for planning tests}
\label{sim-forest}
\begin{center}
\renewcommand{\arraystretch}{1.25} 
\begin{tabular}{c c c c c}
\toprule
Env. & Algorithm & AR (\%) & APL (m) & SPL (\%)\\
\midrule
\multirow{3}{*}{Forest} & Bug (left) & 99.7 & 21.328 & 87.3\\
 & Bug (right) & 98.7 & 21.320 & 87.3\\
 & \textbf{Proposed} & \textbf{100} & \textbf{19.452} & \textbf{95.7}\\
\hline
\multirow{3}{*}{Rocky} & Bug (left) & 99.3 & 20.780 & 90.1\\
 & Bug (right) & 100 & 20.688 & 90.5\\
 & \textbf{Proposed} & \textbf{100} & \textbf{19.197} & \textbf{97.6}\\
\bottomrule
\end{tabular}
\end{center}
\end{table}

\section{Conclusion}
\label{sec:Conclusion}
In this paper, we proposed a navigation method for robot swarms based on FFT filtering and probabilistic fusion decision. We designed two types of filters according to the robot’s physical dimensions for extracting safe directions and compression of environmental observations, respectively. Then, the observation fusion and decision were achieved in the probabilistic domain. By conducting a combination of simulated and real-world experiments, we demonstrated the effectiveness of our proposed method in complex unknown environments, as well as the very low computational and communication costs. Our approach is capable of processing 3D environmental observation data. Future work includes improvements of filters and performance tests for 3D application scenarios.

\bibliography{IEEEabrv,egbib.bib}
\end{document}